\documentclass[10pt,twocolumn,letterpaper]{article}
\usepackage[letterpaper,top=1in,bottom=1in,left=1in,right=1in]{geometry}
\usepackage[T1]{fontenc}
\usepackage{times}
\usepackage{amsmath,amssymb}
\usepackage{graphicx}
\usepackage{booktabs}
\usepackage{multirow}
\usepackage{caption}
\usepackage{titlesec}
\usepackage[hidelinks]{hyperref}

\titleformat{\section}{\normalfont\large\bfseries}{\thesection.}{0.6em}{}
\titlespacing*{\section}{0pt}{1.6ex plus .4ex minus .2ex}{0.9ex}
\titleformat{\subsection}{\normalfont\normalsize\bfseries}{\thesubsection.}{0.5em}{}
\titlespacing*{\subsection}{0pt}{1.2ex plus .3ex minus .2ex}{0.6ex}

\renewenvironment{abstract}{%
  \par\vspace{0.5em}%
  \begin{center}{\bfseries Abstract}\end{center}%
  \par\begingroup\itshape\noindent\ignorespaces
}{%
  \par\endgroup\vspace{0.9em}%
}

\begin{document}

\twocolumn[{
\begin{center}
{\LARGE\bfseries Input-Space versus Feature-Level Conditioning\\
for Robust Facial Expression Recognition\par}
\vspace{1.0em}
{\large Weihang Yuan\par}
\vspace{0.35em}
{\normalsize \texttt{yuanweihang810@gmail.com}\par}
\end{center}
\vspace{0.3em}
}]

\begin{abstract}
Facial expression recognition (FER) degrades sharply when illumination, exposure or resolution departs from the training distribution. Two remedies dominate practice: enriching the input signal, or conditioning intermediate features on an estimated image quality. The two are rarely compared under matched backbones, training budgets and condition annotations, and their cost is seldom reported. We present a controlled comparison on RAF-DB. Input-space conditioning uses four deterministic, label-free views---logarithmic stretching, inverse-logarithmic stretching, down--up resampling and identity---each processed by a shared backbone with a branch-specific BatchNorm and linear head, with the branch logits averaged. Feature-space conditioning is implemented as FiLM and Conditional BatchNorm under an identical protocol. Across six backbone architectures the input-space variant improves top-1 accuracy on every one (mean $+0.94$ pp; bootstrap interval $[0.51,1.63]$, sign test $p=0.016$), with a seed-controlled estimate of $+0.42$ pp on ResNet-18, whereas feature-level conditioning yields $+0.10$ and $+0.03$ pp. We attribute the gap to the information content of the heuristic condition label, not to the conditioning mechanism. The gain costs $4.00\times$ FLOPs, $4.18\times$ latency and $3.79\times$ training time, i.e.\ $0.26$ pp per additional GFLOP, and we state an explicit rule for when the trade is worth making.
\end{abstract}

\section{Introduction}
Facial expression recognition (FER) is a core task in affective computing and is increasingly deployed in clinical and behavioural monitoring, including pain assessment, depression screening and assistive technologies for autism spectrum disorder. Deep networks reach high accuracy on curated benchmarks such as RAF-DB \cite{li2017reliable}, AffectNet \cite{mollahosseini2019affectnet} and FER2013 \cite{goodfellow2013challenges}; a recent survey covers the architectural and loss-design landscape \cite{li2022survey}. That accuracy is measured on images whose acquisition conditions are close to the training distribution. In deployment the same expression may be recorded under dim ward lighting, under saturated clinical illumination, or with a low-resolution camera, and recognition accuracy drops accordingly \cite{dodge2016understanding,hendrycks2019benchmarking}.

Two families of remedies appear in the literature. The first keeps the network fixed and enriches the input distribution, typically through photometric and geometric augmentation \cite{shorten2019survey}. The second keeps the input fixed and modulates the network, either through an explicit quality estimate or through feature-level conditioning operators such as FiLM \cite{perez2018film} or Conditional BatchNorm \cite{devries2017modulating}. The families embody different assumptions: augmentation assumes that a single representation can be made invariant to the nuisance factor if it is seen often enough; feature conditioning assumes that the nuisance factor can be estimated at inference time and used to re-parameterise the representation.

Both assumptions are questionable for image degradation. Contrast loss is not a nuisance that leaves the class signal intact: it destroys information in specific intensity bands, and the destroyed band differs between under- and over-exposure. Feature conditioning, in turn, depends on the quality of the condition estimate; when that estimate is itself derived from a heuristic rule, the modulation may carry little information.

This paper asks a narrow, testable question: \emph{at a fixed backbone, fixed training budget and fixed condition annotation, does conditioning the input signal outperform conditioning the features?} We answer it with a controlled study rather than a new architecture. Our input-space variant applies four fixed, label-free transformations to the input---a logarithmic stretch matched to low illumination, an inverse-logarithmic stretch matched to over-exposure, a down--up resampling matched to low resolution, and an identity mapping---passes each view through a shared backbone with a branch-specific BatchNorm and linear head, and averages the resulting logits. Our feature-space baselines are FiLM and Conditional BatchNorm with the same backbone, optimiser, schedule and annotation.

The contribution is fourfold. (i) A controlled, like-for-like comparison of input-space and feature-space conditioning in which every factor except the locus of conditioning is held fixed. (ii) A six-backbone replication showing that the input-space gain is not specific to one architecture, with an explicit statement of how much of that gain rests on a single backbone. (iii) A seed-controlled estimate of the effect size, reported alongside the single-run numbers rather than instead of them. (iv) A complete cost account---parameters, FLOPs, inference latency and training time---from which we derive a simple decision rule for deployment. We also report a negative result that we consider the most useful part of the study: under weak, rule-based condition annotation, feature-level conditioning provides essentially no benefit, whereas label-free input transformations do. This reframes the choice between the two families as a choice about what supervision is actually available.

\section{Related Work}
\subsection{FER under Degraded Imaging}
Deep FER has moved from hand-crafted descriptors to convolutional backbones \cite{he2016deep} with channel and spatial attention \cite{hu2018squeeze,woo2018cbam}, and more recently to transformer encoders \cite{dosovitskiy2021image}. Representative lines address label noise and annotation uncertainty \cite{wang2020suppressing,zhang2021relative}, part-based and region-aware representations \cite{wang2020region}, feature decomposition \cite{ruan2021feature}, and relation-aware transformers \cite{xue2021transfer}. These methods generally assume that the input has already been captured under acceptable conditions.

The effect of acquisition quality has been studied separately. Dodge and Karam \cite{dodge2016understanding} showed that mild blur, noise and contrast changes produce large accuracy losses in ImageNet-trained networks, and Hendrycks and Dietterich \cite{hendrycks2019benchmarking} established the corruption benchmarks now standard for robustness claims. Augmentation-based remedies such as AugMix \cite{hendrycks2020augmix} and training on Gaussian-noise-corrupted images \cite{rusak2020simple} improve robustness without changing the architecture, and are the closest precedent for the input-space approach studied here. Work on texture bias \cite{geirhos2019texture} offers a partial explanation: convolutional networks rely heavily on high-frequency texture, exactly the band that blur and downsampling remove. Our low-resolution view is motivated by this observation.

For context, published RAF-DB accuracies range from $86.9\%$ (region attention networks \cite{wang2020region}) and $87.0\%$ (self-cure networks \cite{wang2020suppressing}) to $90.9\%$ (TransFER \cite{xue2021transfer}). Those figures are not directly comparable with ours: the strongest entries rely on face-specific pretraining, landmark supervision or transformer backbones an order of magnitude larger, whereas every model here is an ImageNet-initialised generic backbone trained through one identical pipeline. The quantity we measure is the \emph{difference} between two conditioning strategies at a fixed backbone, not the absolute position on that leaderboard.

\subsection{Quality-Aware and Multi-Branch Models}
Quality-aware learning addresses degradation through multi-branch architectures, enhancement front-ends, or domain adaptation. These designs share a common structure: several processing paths followed by a fusion step, superficially related to mixture-of-experts models \cite{jacobs1991adaptive,shazeer2017outrageously} in which a learned gate routes each input to specialised sub-networks. The difference is that our branches share one backbone and are paired with views a priori; we make no claim that the branches learn a clean one-to-one specialisation, and we treat that as an open question.

\subsection{Feature-Level Conditioning and Multi-View Aggregation}
FiLM \cite{perez2018film} applies a learned affine transformation to intermediate feature maps, with the affine parameters produced from a conditioning input; Conditional BatchNorm \cite{devries2017modulating} replaces the global normalisation statistics with condition-specific ones. Both add few parameters and can be inserted into an existing backbone, but their success has been demonstrated mainly in visual question answering and image generation, where the conditioning signal is a clean, well-defined modality. Whether they transfer to a condition label obtained by thresholding image statistics has, to our knowledge, not been isolated. Multi-view aggregation, by contrast, combines predictions from several representations by majority vote, weighted averaging or learned attention; test-time augmentation is the limiting case in which the views are label-preserving and the aggregation is fixed. Our method is structurally close to TTA with four deterministic views, but the views are not all label-preserving in the same sense: the logarithmic stretches are designed to expose information in different intensity bands rather than to average out nuisance.

\section{Method}
\subsection{Problem Setting}
Let $x \in [0,1]^{3 \times H \times W}$ be an aligned face image and $y \in \{1,\dots,C\}$ its expression label, with $C=7$ for RAF-DB. The acquisition condition is not observed; at most a heuristic condition label $c$ is available, and in our setting it is available at both training and test time. Conditioning can be inserted on the input, producing views $\{T_k(x)\}_{k=1}^{K}$, or on the features, producing $\{h_k(f_\theta(x))\}_{k=1}^{K}$ where $h_k$ is a condition-dependent modulation.

\subsection{Input-Space Views}
We use $K=4$ deterministic, parameter-free transformations, each matched to one degradation family. For low illumination, a logarithmic stretch
\begin{equation}
T_1(x) = \frac{\log(1+\lambda x)}{\log(1+\lambda)}, \qquad \lambda = 20,
\end{equation}
expands the low-intensity range and compresses the high-intensity range, so that structure in the dark end of the histogram becomes visible; this is the standard log transform of image processing \cite{gonzalez2018digital}. For over-exposure, the complementary inverse-logarithmic stretch
\begin{equation}
T_2(x) = 1 - \frac{\log(1+\lambda(1-x))}{\log(1+\lambda)}, \qquad \lambda = 20,
\end{equation}
expands the high-intensity range instead. For low resolution, down--up resampling
\begin{equation}
T_3(x) = \mathrm{up}\big(\mathrm{down}(x,s),s^{-1}\big), \qquad s = 0.3,
\end{equation}
with bilinear interpolation for both operators, removes high-frequency detail and forces reliance on coarse shape, consistent with texture-bias observations \cite{geirhos2019texture}. The fourth view is the identity $T_4(x)=x$, which preserves the original statistics and prevents the transformed views from dominating the average.

More elaborate contrast operators exist, notably contrast-limited adaptive histogram equalisation (CLAHE) \cite{zuiderveld1994contrast} and multiscale Retinex \cite{jobson1997multiscale}; we deliberately use the simplest parameter-free point operations so that the comparison isolates the effect of having several views rather than the effect of a tuned enhancement algorithm. The two hyper-parameters were selected by a grid search on the validation split over $\lambda \in \{5,10,20,30\}$ and $s \in \{0.2,0.3,0.5\}$; we did not observe a sharp optimum, and the reported values are the centre of a broad plateau.

\subsection{Multi-View Architecture}
Each view is processed by a shared backbone $f_\theta$ followed by a branch-specific normalisation and classifier:
\begin{equation}
\begin{aligned}
v_k &= T_k(x), \quad z_k = f_\theta(v_k),\\
u_k &= \mathrm{BN}_k(z_k), \quad \ell_k = g_k(u_k) \in \mathbb{R}^{C}.
\end{aligned}
\end{equation}
The backbone is shared, so the additional parameters are one BatchNorm layer and one linear head per branch; the overhead is stated exactly per backbone in Table~\ref{tab:eff}.

\begin{figure*}[t]
\centering
\includegraphics[width=0.92\textwidth]{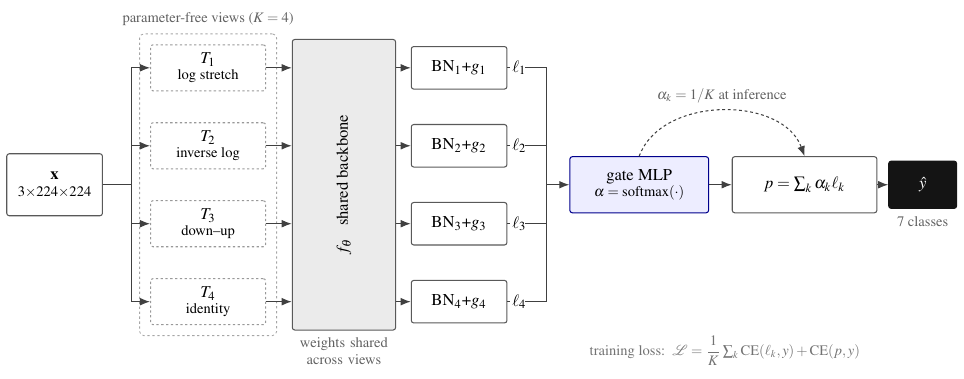}
\caption{Overview of the multi-view pipeline. Four parameter-free transformations (left) map the input to condition-specific views; each view passes through the same backbone, whose weights are shared; only the BatchNorm and linear head are branch-specific. A small gate network produces the aggregation weights $\alpha_k$, and the weighted sum of the branch logits is the prediction. At inference the gate can be replaced by the uniform choice $\alpha_k=1/K$ without measurable loss of accuracy.}
\label{fig:arch}
\end{figure*}

The final logits are a weighted combination
\begin{equation}
p = \sum_{k=1}^{K} \alpha_k \ell_k, \qquad \sum_k \alpha_k = 1,
\end{equation}
where $\alpha_k$ is produced by a small gating network applied to the concatenated logits (a two-layer perceptron with hidden width 32). At inference the learned weights can be replaced by the uniform choice $\alpha_k = 1/K$; on ResNet-18 this changes test accuracy from $86.21\%$ to $86.25\%$, a difference far below the seed-to-seed standard deviation of $0.23$--$0.28$ pp. The gate is therefore close to uniform and can be removed at deployment at no cost in accuracy, saving $1{,}952$ parameters. Because $p$ is a weighted average of sigmoids it is a score rather than a normalised distribution; we use it for ranking only and do not interpret it as a calibrated probability \cite{guo2017calibration}.

Training minimises the sum of a per-branch cross-entropy and a cross-entropy on the aggregated prediction,
\begin{equation}
\mathcal{L} = \frac{1}{K}\sum_{k=1}^{K} \mathrm{CE}(\ell_k, y) + \mathrm{CE}(p, y),
\end{equation}
with class weights inversely proportional to class frequency in both terms. The second term couples the branches through the gate; without it the gate receives no direct gradient.

\subsection{Scope of the Claim}
The word \emph{condition-specific} is a statement about design, not about adaptation. Each view is matched to a degradation family a priori; the method contains no condition classifier and no per-image view selection. We therefore do not claim that the network learns to route an image to the branch that matches its condition. What we claim is weaker and testable: exposing the network to a fixed set of degradation-matched views, and averaging their predictions, is a better use of a fixed backbone than modulating its features with a heuristic condition label.

\section{Experimental Protocol}
\subsection{Dataset and Training}
All experiments use RAF-DB \cite{li2017reliable}, which contains $12{,}271$ training and $3{,}068$ test images annotated with seven basic expressions. Ten percent of the training set is held out for validation and early stopping, stratified by class; the test set is touched once per configuration. Images are aligned with MTCNN \cite{zhang2016joint} and resized to $224\times224$. We use AdamW \cite{loshchilov2019decoupled} with learning rate $5\times10^{-4}$, weight decay $5\times10^{-3}$ and a cosine schedule \cite{loshchilov2017sgdr}. Batch size is 64, reduced to 48 for ResNet-50 and DenseNet-121 for memory reasons. Training runs for 60 epochs, except ResNet-18 with four views, which runs for 80 epochs because it has not converged at 60; in that configuration the single-view baseline is also trained for 80 epochs so that the comparison remains budget-matched. Early stopping uses validation accuracy with patience 15 (30 for the 80-epoch configuration). Augmentation is limited to random horizontal flip, rotation within $\pm 10^\circ$, and colour jitter \cite{shorten2019survey}. All models are initialised from ImageNet weights \cite{deng2009imagenet}.

\subsection{Baselines and Condition Annotation}
Five baselines are used: (1) a single-view network ($K{=}1$, identity view) under the same protocol; (2) a random-geometric-view variant ($K{=}4$) in which the four views are rotations of $0^\circ$, $90^\circ$, $180^\circ$ and $270^\circ$, which isolates condition alignment from generic multi-view ensembling; (3) FiLM \cite{perez2018film}, applying a learned affine modulation to the pooled features; (4) Conditional BatchNorm \cite{devries2017modulating}; (5) a single-view ResNet-50 with horizontal-flip TTA, which has approximately the same FLOP budget as the four-view ResNet-18 ($8.26$ vs $7.29$ GFLOPs) and therefore tests whether the gain can be bought more cheaply by a larger model with TTA. A single-view ResNet-101 was trained as a capacity control.

The conditioning baselines require a condition label. We use a three-class annotation derived from simple intensity statistics---a dark, a bright and a standard class---computable at test time and therefore not assuming unavailable supervision. Its distribution over RAF-DB ($n=15{,}339$) is strongly imbalanced: dark $10.9\%$ ($1{,}671$), bright $53.2\%$ ($8{,}165$), standard $35.9\%$ ($5{,}503$). A modulation conditioned on such a label has, in effect, one dominant regime and two minority regimes, and the minority regimes are precisely the ones for which conditioning should matter.

\subsection{Statistical Analysis and Timing}
We report three quantities separately rather than pooling them into a headline number: the mean gain over the six backbones with a descriptive bootstrap interval obtained by resampling the six backbone-level differences $20{,}000$ times (with $n=6$ this characterises the spread across architectures, not sampling error); a sign test on the direction across backbones, where six positive differences out of six occur with probability $0.5^{6}=0.016$ under the null; and, for ResNet-18, paired differences over three seeds ($42$, $123$, $2024$), reported as mean $\pm$ standard deviation with the paired $t$ statistic. With three seeds the test has low power, and we interpret a non-significant result as uninformative rather than as evidence of no effect. Parameter counts and FLOPs are computed on the model graph at $224\times224$. Inference latency is the median of 100 forward passes at batch size 1 on a single GPU after 10 warm-up iterations; training time is the median wall-clock time per optimisation step at the training batch size. Both are reported as ratios to the single-view baseline.

\section{Results}
\subsection{Main Results across Backbones}
Table~\ref{tab:main} reports test accuracy for six backbones. The multi-view model is better than its single-view counterpart on all six. The mean gain is $+0.94$ pp, with a descriptive bootstrap interval of $[0.51,1.63]$ and a sign-test $p$ of $0.016$.

\begin{figure}[t]
\centering
\includegraphics[width=0.98\columnwidth]{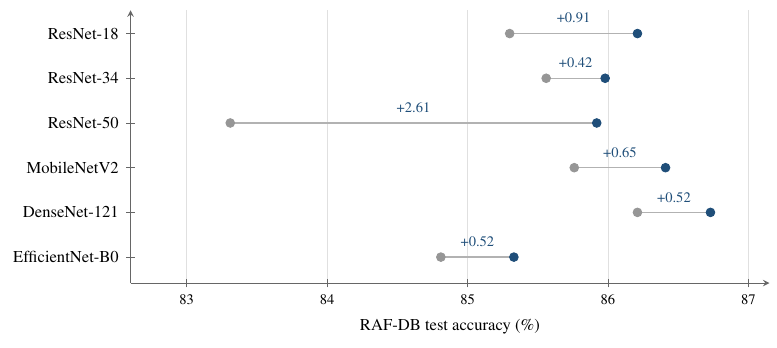}
\caption{RAF-DB test accuracy for the six backbones. Grey dots are the single-view baselines, blue dots the multi-view model, and the label on each segment gives the gain in percentage points. The accuracy axis is truncated; absolute values are listed in Table~\ref{tab:main}.}
\label{fig:main}
\end{figure}

\begin{table}[t]
\centering
\caption{RAF-DB test accuracy (\%) on six backbones. Single run per configuration; best per row in bold.}
\label{tab:main}
\footnotesize
\setlength{\tabcolsep}{4pt}
\renewcommand{\arraystretch}{1.1}
\begin{tabular*}{\columnwidth}{@{\extracolsep{\fill}}|l c c c|}
\hline
Backbone & Single ($K{=}1$) & Multi-view ($K{=}4$) & $\Delta$\\
\hline
ResNet-18       & 85.30 & \textbf{86.21} & $+0.91$\\
ResNet-34       & 85.56 & \textbf{85.98} & $+0.42$\\
ResNet-50       & 83.31 & \textbf{85.92} & $+2.61$\\
MobileNetV2     & 85.76 & \textbf{86.41} & $+0.65$\\
DenseNet-121    & 86.21 & \textbf{86.73} & $+0.52$\\
EfficientNet-B0 & 84.81 & \textbf{85.33} & $+0.52$\\
\hline
Mean            & 85.16 & \textbf{86.10} & $\mathbf{+0.94}$\\
\hline
\end{tabular*}
\end{table}

Two features of Table~\ref{tab:main} bound the strength of the conclusion. First, the gain is unevenly distributed: ResNet-50 contributes $+2.61$ pp, which is $46\%$ of the summed gain, while ResNet-34 contributes $+0.42$ pp. The median gain is $0.59$ pp and the mean excluding ResNet-50 is $+0.60$ pp; a pooled mean over six architectures therefore overstates the typical effect, and we report it as a summary rather than as an estimate of what a new backbone would gain. Second, the ResNet-50 single-view baseline at $83.31\%$ is lower than our other baselines, which fall between $84.81\%$ and $86.21\%$; we therefore report the median and the seed-controlled estimate alongside the pooled mean, and we treat the ResNet-50 row as an upper bound on the benefit for that architecture rather than as the headline result.

\subsection{Seed-Controlled Comparison}
Table~\ref{tab:seed} reports a three-seed study on ResNet-18 under a unified 60-epoch protocol, deliberately more conservative than the 80-epoch configuration of Table~\ref{tab:main}. The multi-view model is better than the single-view baseline on all three seeds and better than the random-view variant on all three seeds.

\begin{table}[t]
\centering
\caption{Three-seed ResNet-18 results on RAF-DB (mean $\pm$ std, \%). The random-view row replaces condition-specific views with rotations of $0^\circ$, $90^\circ$, $180^\circ$ and $270^\circ$.}
\label{tab:seed}
\footnotesize
\setlength{\tabcolsep}{3pt}
\renewcommand{\arraystretch}{1.1}
\begin{tabular*}{\columnwidth}{@{\extracolsep{\fill}}|l c c c c|}
\hline
Method & 42 & 123 & 2024 & Mean\\
\hline
Single view     & 85.63 & 85.07 & 85.43 & $85.38 \pm 0.28$\\
Random views    & 85.33 & 85.46 & 85.37 & $85.39 \pm 0.07$\\
Condition views & 85.82 & 86.02 & 85.56 & $\mathbf{85.80 \pm 0.23}$\\
\hline
$\Delta$ vs single  & $+0.19$ & $+0.95$ & $+0.13$ & $+0.42 \pm 0.46$\\
$\Delta$ vs random  & $+0.49$ & $+0.56$ & $+0.19$ & $+0.41 \pm 0.20$\\
\hline
\end{tabular*}
\end{table}

The paired differences against the single-view baseline are $+0.19$, $+0.95$ and $+0.13$ pp, with mean $+0.42$ pp and standard deviation $0.46$ pp ($t=1.60$, $n=3$). With three seeds this test has low power, and the honest reading is that the direction of the effect is stable while its magnitude is not: the paired gain ranges from $+0.13$ to $+0.95$ pp. The comparison with random views is more informative. The paired differences are $+0.49$, $+0.56$ and $+0.19$ pp (mean $+0.41$, $t=3.64$), and the random-view model is indistinguishable from the single-view baseline ($85.39$ vs $85.38$), which confirms that adding four branches with arbitrary geometric views does not by itself produce a gain. What separates the two four-view models is the choice of views, not the number of branches: the effect is attributable to condition alignment rather than to ensembling. The gap between $+0.42$ pp here and $+0.91$ pp for ResNet-18 in Table~\ref{tab:main} is explained by the training budget: the 80-epoch configuration is $0.41$ pp better than the 60-epoch one, close to the seed-to-seed standard deviation. We report both and do not present the larger as the effect size.

\subsection{Input-Space versus Feature-Space Conditioning}
Table~\ref{tab:cond} compares the two loci of conditioning under an identical backbone, optimiser, schedule and annotation. Feature-level conditioning adds $0.10$ pp (FiLM) and $0.03$ pp (Conditional BatchNorm), both an order of magnitude below the input-space gain and both below the seed-to-seed standard deviation of the backbone.

\begin{table}[t]
\centering
\caption{Input-space versus feature-space conditioning on RAF-DB (ResNet-18), same three-class annotation throughout.}
\label{tab:cond}
\footnotesize
\setlength{\tabcolsep}{4pt}
\renewcommand{\arraystretch}{1.1}
\begin{tabular*}{\columnwidth}{@{\extracolsep{\fill}}|l c c|}
\hline
Method & Accuracy (\%) & $\Delta$ vs single\\
\hline
Single view ($K{=}1$) & 85.27 & ---\\
FiLM \cite{perez2018film} & 85.37 & $+0.10$\\
Conditional BatchNorm \cite{devries2017modulating} & 85.30 & $+0.03$\\
Input-space views ($K{=}4$, ours) & \textbf{86.21} & $\mathbf{+0.94}$\\
\hline
\end{tabular*}
\end{table}

We do not interpret this as evidence that FiLM or Conditional BatchNorm are ineffective in general. Both operators are functions of the condition label; if that label carries little information about the actual degradation, the modulation cannot compensate for it. Our annotation is a three-class thresholding of two global image statistics: it separates bright from standard images reasonably well but places only $10.9\%$ of the data in the dark class, and a single global threshold cannot distinguish a uniformly dark image from one with a bright background and a dark face. Under such a label, the affine parameters learned for the dark class are estimated from a small and heterogeneous subset. The input-space view requires no label at all: the logarithmic stretch is applied to every image, and the network can ignore it when it is unhelpful.

\subsection{Gains on Naturally Degraded Subsets}
Partitioning the test set by image statistics, the full test set gains $+0.94$ pp ($3{,}068$ images, $\approx$29 images), the bottom third by mean brightness gains $+0.78$ pp ($1{,}023$ images, $\approx$8), and the bottom third by Laplacian variance gains $+1.37$ pp ($1{,}023$ images, $\approx$14). Only the low-resolution subset shows a gain larger than the full-set gain, which is the opposite of what a naive reading of the method predicts, since the logarithmic stretch is designed for dark images. Two explanations are compatible with the data: the brightness threshold may select images that are dark but not necessarily under-exposed, or the subset may simply be too small, since a difference of $0.78$ pp on $1{,}023$ images corresponds to about eight images. Under either explanation the subset result is weak supporting evidence, not a confirmation of the design rationale.

\subsection{Computational Cost}
Table~\ref{tab:eff} reports parameters, FLOPs, latency and training time. The parameter overhead is small but not uniformly $0.2\%$: that figure holds for ResNet-18 ($11.18\to11.20$ M) and is the smallest in the table; it reaches $1.34\%$ for MobileNetV2, where the backbone itself is small, and averages $0.49\%$. The computational overhead is close to $4\times$ everywhere, as expected from four forward passes: $4.00\times$ FLOPs, $4.18\times$ latency and $3.79\times$ training time. Normalising the accuracy gain by the additional compute gives a mean of $0.26$ pp per extra GFLOP, with a range from $0.04$ (ResNet-34) to $0.67$ (MobileNetV2).

\begin{table*}[t]
\centering
\caption{Cost of the multi-view model relative to the single-view baseline. Latency at batch size 1 on one GPU; training time per optimisation step.}
\label{tab:eff}
\small
\setlength{\tabcolsep}{6pt}
\renewcommand{\arraystretch}{1.15}
\begin{tabular*}{\textwidth}{@{\extracolsep{\fill}}|l c c c c c c c|}
\hline
\multirow{2}{*}{Backbone} & \multicolumn{2}{c}{Params (M)} & \multicolumn{2}{c}{FLOPs (G)} & \multicolumn{2}{c}{Latency (ms)} & Train\\
\cline{2-3}\cline{4-5}\cline{6-7}\cline{8-8}
 & Base & Ours & Base & Ours & Base & Ours & ratio\\
\hline
ResNet-18       & 11.18 & 11.20 & 1.82 & 7.29  & 1.11 & 4.66  & $3.91\times$\\
ResNet-34       & 21.29 & 21.30 & 3.68 & 14.71 & 1.91 & 8.90  & $3.85\times$\\
ResNet-50       & 23.53 & 23.58 & 4.13 & 16.53 & 2.58 & 10.63 & $3.89\times$\\
MobileNetV2     & 2.24  & 2.27  & 0.33 & 1.30  & 2.14 & 9.96  & $4.01\times$\\
DenseNet-121    & 6.96  & 6.99  & 2.90 & 11.58 & 7.22 & 26.95 & $3.11\times$\\
EfficientNet-B0 & 4.02  & 4.05  & 0.41 & 1.66  & 3.65 & 13.63 & $3.98\times$\\
\hline
Mean & --- & $+0.49\%$ & --- & $4.00\times$ & --- & $4.18\times$ & $3.79\times$\\
\hline
\end{tabular*}
\end{table*}

Two baselines test whether the same budget could be spent more efficiently. A single-view ResNet-101 ($44.5$ M parameters, $7.8$ GFLOPs) reached only $80$--$81\%$ validation accuracy under the same protocol, indicating that scaling capacity alone overfits the $12{,}271$ training images. A single-view ResNet-50 with horizontal-flip TTA ($8.26$ GFLOPs) reached $83.80\%$, below the four-view ResNet-18 ($86.21\%$ at $7.29$ GFLOPs). Neither substitute reproduces the gain, which supports the interpretation that the benefit comes from the diversity of the views rather than from the extra compute per se.

\section{Discussion}
\subsection{Why Input-Space Conditioning Helps, and Why Only a Little}
The stretches redistribute the intensity histogram so that a different band occupies the linear response range of the first convolutional layers: for an under-exposed image the dark band contains most of the facial structure and the stretch makes it representable, and for an over-exposed image the same is true of the bright band. The resampling view removes high-frequency texture, which the literature identifies as a dominant cue in convolutional networks \cite{geirhos2019texture}, and forces reliance on coarser shape. Because the backbone is shared, the four views regularise it toward features that survive all four transforms---in effect a form of consistency training implemented at the input.

The small size of the effect is equally explicable: RAF-DB is a curated, web-collected dataset in which severe degradation is rare, so most test images are already well exposed and adequately resolved. A method designed for degraded inputs has little room to improve on images that are not degraded, and the full-set gain is therefore diluted. This is consistent with the low-resolution subset showing the largest gain and with the near-zero gain of the random-view control, which confirms that the improvement is not a generic ensemble effect.

\subsection{Why Feature-Level Conditioning Underperforms Here}
Three factors specific to our setting plausibly account for Table~\ref{tab:cond}. In the terminology of shortcut learning \cite{geirhos2020shortcut}, a label that is a deterministic function of the acquisition process invites the network to identify the process rather than to compensate for it; our condition label is a global threshold on two statistics and cannot represent spatially varying illumination, which is common in unconstrained face images. The label distribution is strongly imbalanced, so the parameters for the minority class are estimated from few examples. And the modulation is applied at a single point in the network, whereas the information destroyed by clipping or downsampling is destroyed before the first layer and cannot be restored downstream.

The last point is the substantive one. Modulation re-weights what the network already represents; it does not create information that the input no longer contains. An input-space transform, by contrast, changes what the first layer sees. If the degradation is a loss of dynamic range, only the second kind of intervention can recover the lost band---consistent with the ordering in Table~\ref{tab:cond}, and suggesting that the comparison would shift in favour of feature conditioning only if the condition signal were richer than the image statistics we used.

\subsection{When the Cost Is Justified}
The intervention multiplies inference cost by about $4.2$ and training cost by about $3.8$, and returns $0.26$ pp per additional GFLOP on average, with a seed-controlled estimate of $+0.42$ pp on the reference backbone. Three regimes follow. For \emph{offline analysis}, where throughput is not binding, the cost is irrelevant and the method can be applied whenever a sub-percentage-point gain matters. For \emph{interactive or near-real-time use}, the $4.2\times$ latency is tolerable only if the baseline already runs well above the required frame rate: ResNet-18 at $1.11$ ms per image leaves ample headroom at $4.66$ ms, whereas MobileNetV2 at $2.14$ ms to $9.96$ ms is borderline for a $30$ fps budget shared with other components. For \emph{embedded or battery-powered monitoring}, the method is not appropriate in its current form: a single forward pass of the smallest backbone is already the dominant cost, and multiplying it by four is difficult to justify for less than one percentage point. Pruning the view set, or distilling the four-view model into a single-view student, is the natural remedy and is left for future work.

\subsection{Limitations}
\emph{Single dataset.} All results are on RAF-DB, a curated dataset whose degradations are mild; the conclusions may not transfer to genuinely low-quality acquisition, such as patient monitoring video. A second dataset with controlled, physically motivated degradations is the single most important extension. \emph{Seed coverage.} Only ResNet-18 has three seeds; the other five backbones are single runs, so the pooled mean mixes architectures with different variances, and the seed-controlled estimate of $+0.42$ pp should be preferred when a single effect size is required. \emph{Condition annotation.} The three-class annotation is heuristic and imbalanced, which weakens the feature-conditioning baselines; a learned quality estimator, or a continuous quality score, would provide a fairer test.

\section{Conclusion}
We compared two loci for conditioning a facial expression recogniser against image degradation, holding backbone, schedule and condition annotation fixed. Conditioning the input with four fixed, label-free views improved accuracy on all six backbones tested, with a pooled mean of $+0.94$ pp (bootstrap interval $[0.51,1.63]$, sign test $p=0.016$) and a seed-controlled estimate of $+0.42$ pp on ResNet-18. Conditioning the features with FiLM or Conditional BatchNorm, using the same annotation, produced $+0.10$ and $+0.03$ pp. The gap is best explained by the information content of the condition label rather than by the conditioning mechanism: a coarse, imbalanced label cannot tell the network which intensity band has been lost, whereas an input-space transform does not need to be told.

The gain is modest and expensive---$4.00\times$ FLOPs, $4.18\times$ latency and $3.79\times$ training time, or $0.26$ pp per additional GFLOP---and we have stated the cost alongside the benefit together with a rule for when the trade is worth making. Three extensions would materially strengthen the work: a second dataset with controlled degradations; a branch-by-condition matrix establishing whether the views specialise; and a distillation or pruning procedure that recovers most of the gain at single-view cost. The last is, in our view, the most useful, because it would convert a $4\times$ method with a small gain into a deployable one.


\begin{thebibliography}{99}
\small

\bibitem{li2017reliable}
S. Li, W. Deng, and J. Du, \textquotedblleft{}Reliable crowdsourcing and deep locality-preserving learning for expression recognition in the wild, \textquotedblright in \emph{Proc. IEEE Conf. Comput. Vis. Pattern Recognit. (CVPR)}, 2017, pp. 2852--2861.

\bibitem{mollahosseini2019affectnet}
A. Mollahosseini, B. Hasani, and M. H. Mahoor, \textquotedblleft{}AffectNet: A database for facial expression, valence, and arousal computing in the wild, \textquotedblright \emph{IEEE Trans. Affect. Comput.}, vol. 10, no. 1, pp. 18--31, 2019.

\bibitem{goodfellow2013challenges}
I. J. Goodfellow \emph{et al.}, \textquotedblleft{}Challenges in representation learning: A report on three machine learning contests, \textquotedblright in \emph{Proc. Int. Conf. Neural Inf. Process. (ICONIP)}, 2013, pp. 117--124.

\bibitem{li2022survey}
S. Li and W. Deng, \textquotedblleft{}Deep facial expression recognition: A survey, \textquotedblright \emph{IEEE Trans. Affect. Comput.}, vol. 13, no. 3, pp. 1195--1215, 2022.

\bibitem{dodge2016understanding}
S. Dodge and L. Karam, \textquotedblleft{}Understanding how image quality affects deep neural networks, \textquotedblright in \emph{Proc. Int. Conf. Quality Multimedia Experience (QoMEX)}, 2016, pp. 1--6.

\bibitem{hendrycks2019benchmarking}
D. Hendrycks and T. Dietterich, \textquotedblleft{}Benchmarking neural network robustness to common corruptions and perturbations, \textquotedblright in \emph{Proc. Int. Conf. Learn. Represent. (ICLR)}, 2019.

\bibitem{shorten2019survey}
C. Shorten and T. M. Khoshgoftaar, \textquotedblleft{}A survey on image data augmentation for deep learning, \textquotedblright \emph{J. Big Data}, vol. 6, no. 1, p. 60, 2019.

\bibitem{perez2018film}
E. Perez, F. Strub, H. de Vries, V. Dumoulin, and A. Courville, \textquotedblleft{}FiLM: Visual reasoning with a general conditioning layer, \textquotedblright in \emph{Proc. AAAI Conf. Artif. Intell.}, 2018, pp. 3942--3951.

\bibitem{devries2017modulating}
H. de Vries, F. Strub, J. Mary, H. Larochelle, O. Pietquin, and A. C. Courville, \textquotedblleft{}Modulating early visual processing by language, \textquotedblright in \emph{Proc. Adv. Neural Inf. Process. Syst. (NeurIPS)}, 2017, pp. 6594--6604.

\bibitem{he2016deep}
K. He, X. Zhang, S. Ren, and J. Sun, \textquotedblleft{}Deep residual learning for image recognition, \textquotedblright in \emph{Proc. IEEE Conf. Comput. Vis. Pattern Recognit. (CVPR)}, 2016, pp. 770--778.

\bibitem{hu2018squeeze}
J. Hu, L. Shen, and G. Sun, \textquotedblleft{}Squeeze-and-excitation networks, \textquotedblright in \emph{Proc. IEEE Conf. Comput. Vis. Pattern Recognit. (CVPR)}, 2018, pp. 7132--7141.

\bibitem{woo2018cbam}
S. Woo, J. Park, J.-Y. Lee, and I. S. Kweon, \textquotedblleft{}CBAM: Convolutional block attention module, \textquotedblright in \emph{Proc. Eur. Conf. Comput. Vis. (ECCV)}, 2018, pp. 3--19.

\bibitem{dosovitskiy2021image}
A. Dosovitskiy \emph{et al.}, \textquotedblleft{}An image is worth 16x16 words: Transformers for image recognition at scale, \textquotedblright in \emph{Proc. Int. Conf. Learn. Represent. (ICLR)}, 2021.

\bibitem{wang2020suppressing}
K. Wang, X. Peng, J. Yang, S. Lu, and Y. Qiao, \textquotedblleft{}Suppressing uncertainties for large-scale facial expression recognition, \textquotedblright in \emph{Proc. IEEE/CVF Conf. Comput. Vis. Pattern Recognit. (CVPR)}, 2020, pp. 6896--6905.

\bibitem{zhang2021relative}
Y. Zhang, C. Wang, and W. Deng, \textquotedblleft{}Relative uncertainty learning for facial expression recognition, \textquotedblright in \emph{Proc. Adv. Neural Inf. Process. Syst. (NeurIPS)}, 2021, pp. 17616--17627.

\bibitem{wang2020region}
K. Wang, X. Peng, J. Yang, D. Meng, and Y. Qiao, \textquotedblleft{}Region attention networks for pose and occlusion robust facial expression recognition, \textquotedblright \emph{IEEE Trans. Image Process.}, vol. 29, pp. 4057--4069, 2020.

\bibitem{ruan2021feature}
D. Ruan, Y. Yan, S. Lai, Z. Chai, C. Shen, and H. Wang, \textquotedblleft{}Feature decomposition and reconstruction learning for effective facial expression recognition, \textquotedblright in \emph{Proc. IEEE/CVF Conf. Comput. Vis. Pattern Recognit. (CVPR)}, 2021, pp. 7656--7665.

\bibitem{xue2021transfer}
F. Xue, Q. Wang, and G. Guo, \textquotedblleft{}TransFER: Learning relation-aware facial expression representations with transformers, \textquotedblright in \emph{Proc. IEEE/CVF Int. Conf. Comput. Vis. (ICCV)}, 2021, pp. 3581--3590.

\bibitem{hendrycks2020augmix}
D. Hendrycks, N. Mu, E. D. Cubuk, B. Zoph, J. Gilmer, and B. Lake, \textquotedblleft{}AugMix: A simple data processing method to improve robustness and uncertainty, \textquotedblright in \emph{Proc. Int. Conf. Learn. Represent. (ICLR)}, 2020.

\bibitem{rusak2020simple}
E. Rusak \emph{et al.}, \textquotedblleft{}A simple way to make neural networks robust against diverse image corruptions, \textquotedblright in \emph{Proc. Eur. Conf. Comput. Vis. (ECCV)}, 2020, pp. 53--69.

\bibitem{geirhos2019texture}
R. Geirhos, P. Rubisch, C. Michaelis, M. Bethge, F. A. Wichmann, and W. Brendel, \textquotedblleft{}ImageNet-trained CNNs are biased towards texture; increasing shape bias improves accuracy and robustness, \textquotedblright in \emph{Proc. Int. Conf. Learn. Represent. (ICLR)}, 2019.

\bibitem{jacobs1991adaptive}
R. A. Jacobs, M. I. Jordan, S. J. Nowlan, and G. E. Hinton, \textquotedblleft{}Adaptive mixtures of local experts, \textquotedblright \emph{Neural Comput.}, vol. 3, no. 1, pp. 79--87, 1991.

\bibitem{shazeer2017outrageously}
N. Shazeer \emph{et al.}, \textquotedblleft{}Outrageously large neural networks: The sparsely-gated mixture-of-experts layer, \textquotedblright in \emph{Proc. Int. Conf. Learn. Represent. (ICLR)}, 2017.

\bibitem{gonzalez2018digital}
R. C. Gonzalez and R. E. Woods, \emph{Digital Image Processing}, 4th ed. New York, NY, USA: Pearson, 2018.

\bibitem{zuiderveld1994contrast}
K. Zuiderveld, \textquotedblleft{}Contrast limited adaptive histogram equalization, \textquotedblright in \emph{Graphics Gems IV}. San Diego, CA, USA: Academic Press, 1994, pp. 474--485.

\bibitem{jobson1997multiscale}
D. J. Jobson, Z. Rahman, and G. A. Woodell, \textquotedblleft{}A multiscale retinex for bridging the gap between color images and the human observation of scenes, \textquotedblright \emph{IEEE Trans. Image Process.}, vol. 6, no. 7, pp. 965--976, 1997.

\bibitem{guo2017calibration}
C. Guo, G. Pleiss, Y. Sun, and K. Q. Weinberger, \textquotedblleft{}On calibration of modern neural networks, \textquotedblright in \emph{Proc. Int. Conf. Mach. Learn. (ICML)}, 2017, pp. 1321--1330.

\bibitem{zhang2016joint}
K. Zhang, Z. Zhang, Z. Li, and Y. Qiao, \textquotedblleft{}Joint face detection and alignment using multitask cascaded convolutional networks, \textquotedblright \emph{IEEE Signal Process. Lett.}, vol. 23, no. 10, pp. 1499--1503, 2016.

\bibitem{loshchilov2019decoupled}
I. Loshchilov and F. Hutter, \textquotedblleft{}Decoupled weight decay regularization, \textquotedblright in \emph{Proc. Int. Conf. Learn. Represent. (ICLR)}, 2019.

\bibitem{loshchilov2017sgdr}
I. Loshchilov and F. Hutter, \textquotedblleft{}SGDR: Stochastic gradient descent with warm restarts, \textquotedblright in \emph{Proc. Int. Conf. Learn. Represent. (ICLR)}, 2017.

\bibitem{deng2009imagenet}
J. Deng, W. Dong, R. Socher, L.-J. Li, K. Li, and L.-F. Fei-Fei, \textquotedblleft{}ImageNet: A large-scale hierarchical image database, \textquotedblright in \emph{Proc. IEEE Conf. Comput. Vis. Pattern Recognit. (CVPR)}, 2009, pp. 248--255.

\bibitem{geirhos2020shortcut}
R. Geirhos \emph{et al.}, \textquotedblleft{}Shortcut learning in deep neural networks, \textquotedblright \emph{Nat. Mach. Intell.}, vol. 2, pp. 665--673, 2020.

\end{thebibliography}
\end{document}